%% file: main_survey_embedded_slam.tex
\documentclass[10pt,final,journal,twoside]{IEEEtran}

\usepackage{blindtext}
\usepackage{amsmath}
\usepackage{hyperref}

\usepackage{graphicx}

\usepackage{url}

\usepackage{array}
\usepackage{multirow}
\usepackage{booktabs}
\usepackage{rotating}

\usepackage{pifont}
\usepackage{xcolor}
\newcommand{\cmark}{\textcolor{green!80!black}{\ding{51}}}

\newif\ifreviews
\newif\ifintro
\newif\ifslampipeline
\newif\ifembedded
\newif\ifconclusion
\newif\ifbiblio

\introtrue 			%Section introduction including: Abstract, Introduction, Related work
\slampipelinetrue		%Section SLAM pipeline from sensors to 3D reconstruction
\embeddedtrue			%Section Real-time processing and low power consumption in embedded systems
\conclusiontrue		%Section Conclusion
\bibliotrue				%Section bibliography and biography

\begin{document}

\title{A survey on real-time 3D scene reconstruction with SLAM methods in embedded systems}

\author{Quentin~Picard,
        Stephane~Chevobbe,
        Mehdi~Darouich,
        and~Jean-Yves Didier% <-this % stops a space
\thanks{Q. Picard, S. Chevobbe and M. Darouich are with CEA, LIST, 91191 Gif-sur-Yvette, France.
(e-mail: quentin.picard@cea.fr)}%
\thanks{Q. Picard and J-Y. Didier are with IBISC, Univ Evry, Universit\'e Paris-Saclay, 91025, Evry, France.}} 
\date{2021}

%\markboth{IEEE TRANSACTIONS ON INTELLIGENT TRANSPORTATION SYSTEMS, VOL. XX, NO. XX, XXXX 2021}{} %First: journal name, Second: author name and paper title
%\markboth{IEEE TRANSACTIONS ON INTELLIGENT TRANSPORTATION SYSTEMS, VOL. XX, NO. XX, XXXX 2021}%
%{Picard \MakeLowercase{\textit{et al.}}: A survey on real-time 3D scene reconstruction with SLAM methods in embedded systems}

\maketitle

\ifreviews
\noindent\textbf{------------------------}
\begin{itemize}
	\item Deadline journal: 11/11/21
	\item 203 references (7 arxiv)
	\item target: 18 pages
	\item 3 figures, 5 tables
	\item \textcolor{blue}{Texte en bleu: }A lire en priorit\'e
\end{itemize}
\textbf{------------------------}\\
\fi

\ifintro
\begin{abstract}
	\input{abstract}
\end{abstract}


%-------------------------------------------------
%-------------------------------------------------
%-------------------------------------------------

\section{Introduction}
	\input{introduction}

%\section{Related work}\label{section:related}
%	\input{relatedwork.tex}
\fi
%-------------------------------------------------
%-------------------------------------------------
%-------------------------------------------------
\ifslampipeline
\section{SLAM pipeline from sensors to 3D reconstruction}\label{section:functions}

\input{figure_main} %FIGURE

	\input{intro_s2.tex}

	\subsection{Localization module}\label{section:localization}
	\input{intro_carto}
        \subsubsection{Front-end}
		    \input{frontend}\label{ssection:functions_frontend}
        	\input{table_techniques} %TABLE
        \subsubsection{Back-end}
            \input{backend}\label{ssection:functions_backend}

    \subsection{3D reconstruction module}\label{section:reconstruction}
		\input{intro_recons}
	    \subsubsection{Mesh reconstruction}
		    \input{mesh}\label{ssection:functions_mesh}

	    \subsubsection{Volumetric reconstruction}
		    \input{volum}\label{ssection:functions_volumetric}
            
	\subsection{Learning-based modules in a conventional pipeline}\label{section:learning}
		\input{intro_dnn}
		\subsubsection{Deep learning for real-time pose estimation}
			\input{dnn_slam}
		\subsubsection{Intermediate representations based on neural networks}
			\input{semantic} 
        \subsubsection{Complexity of deep learning in the context of embedded systems}
			\input{complexity_dnn}

    \subsection{Overview of existing methods}\label{section:benchmark}
        \input{intro_method}
        \subsubsection{Benchmarking tools}
            \input{benchmark}	
        \iffalse
			\subsubsection{Performance evaluation}
            \input{evaluation}
		\fi
		\subsubsection{Performance comparison of existing methods}\label{ssection:methods}
			\input{table_slam}	%TABLE
			\input{table_section3}	%TABLE
			\input{overview_methods.tex}
\fi
%-------------------------------------------------
%-------------------------------------------------
%-------------------------------------------------
\ifembedded
\section{Localization and 3D reconstruction methods on embedded systems}\label{section:hardware}
	\input{intro_s3}
	
	\subsection{Embedded platforms for localization and 3D reconstruction}\label{ssection:sensor_hw}
		\input{intro_hw.tex}
		\subsubsection{Components Off-The-Shelf (COTS)}
			\input{table_fe_hw.tex}	%TABLE
			\input{sensor_hw_cots}
		\subsubsection{Energy-efficient accelerators}
			\input{figure_compare_section}	%FIGURE
			\input{sensor_hw_asi}
		\subsubsection{Vision chips}
			\input{table_section3recons}	%TABLE
			\input{sensor_hw_vchip}
	
	\subsection{Algorithmic methods in embedded platforms}\label{ssection:algo_hw}
		\input{intro_algo_hw.tex}
		\subsubsection{Real-time localization}
			\input{embed_algo_slam}	
		\subsubsection{Real-time 3D reconstruction}
			\input{embed_algo_recons}
\fi
%-------------------------------------------------
%-------------------------------------------------
%-------------------------------------------------
\ifconclusion
\section{Conclusion}\label{section:conclusion}
	\input{conclusion}
\fi

\ifbiblio
\bibliographystyle{IEEEtran}
\bibliography{IEEEabrv,biblio_soa}

\newpage

\begin{IEEEbiography}[{\includegraphics[width=1in,height=1.25in,clip,keepaspectratio]{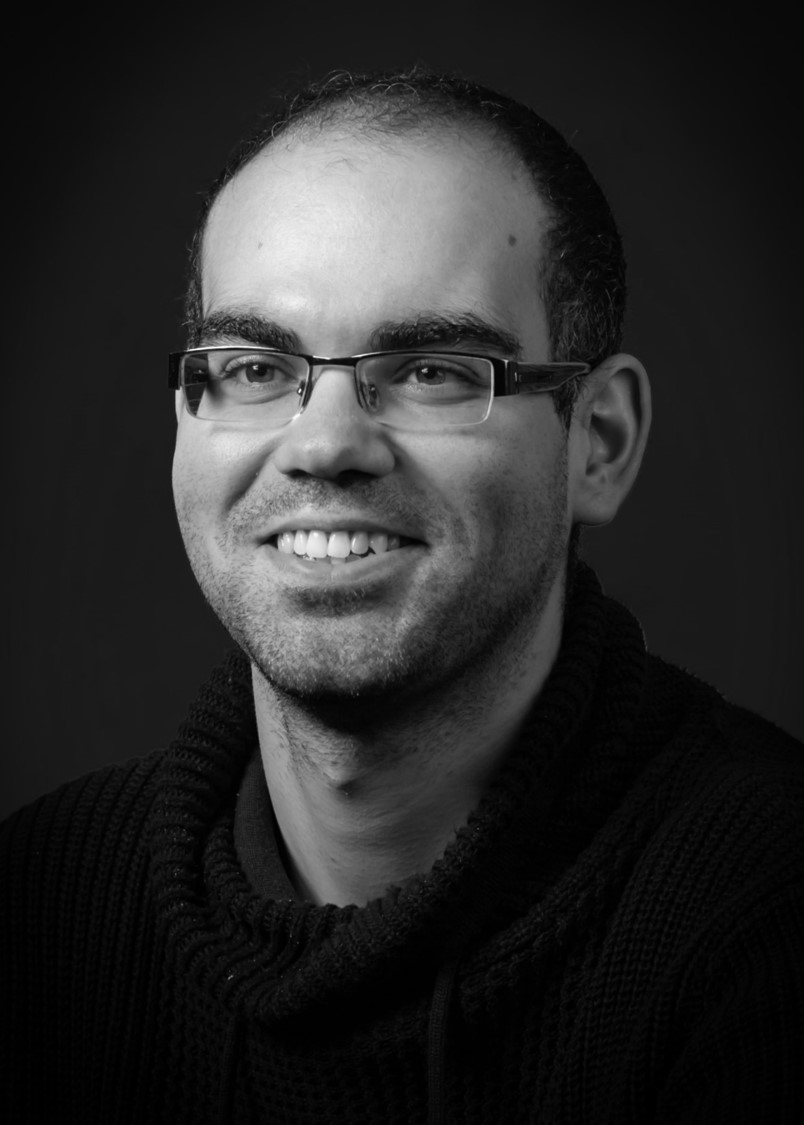}}]{Quentin Picard} 
	received the M.S. degree in mobile autonomous systems from the University of Paris-Saclay, France in
	2019. He is currently pursuing the Ph.D. degree at the CEA LIST Institute, Saclay, France in collaboration with the IBISC
	(Computer Science, Bio-Informatics and Complex Systems) laboratory of the University of Paris-Saclay, France. His research
	interests involve the generation of a semantic and dynamic 3D scene reconstruction in real-time for embedded systems.
\end{IEEEbiography}
\vskip -2\baselineskip plus -1fil
\begin{IEEEbiography}[{\includegraphics[width=1in,height=1.25in,clip,keepaspectratio]{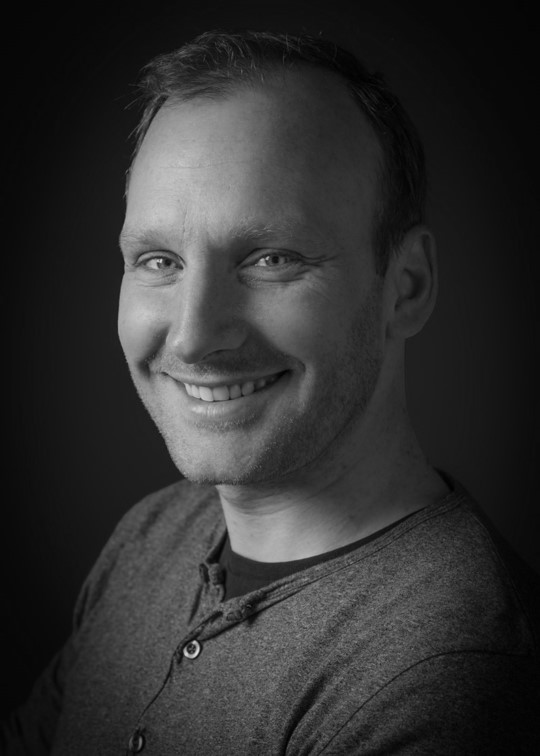}}]{Stephane Chevobbe} 
	received the Ph.D. degree in microelectronic and signal processing from the University of Rennes 1,
	Rennes, France, in 2005. From 2006 to 2009, he participated in several national and European Research Projects that lead to the
	realizations of the application specified integrated circuit and reconfigurable architectures for embedded systems. Since 2009,
	he has participated in the design of computing architectures for embedded vision applications. He is currently an Expert and
	a Research Engineer with the CEA LIST Institute, Saclay, France, where he is involved in the domain of embedded computing
	architecture. His research interests include reconfigurable, programmable and dedicated embedded architectures, and embedded
	architectures for image processing.
\end{IEEEbiography}
\vskip -2\baselineskip plus -1fil
\begin{IEEEbiography}[{\includegraphics[width=1in,height=1.25in,clip,keepaspectratio]{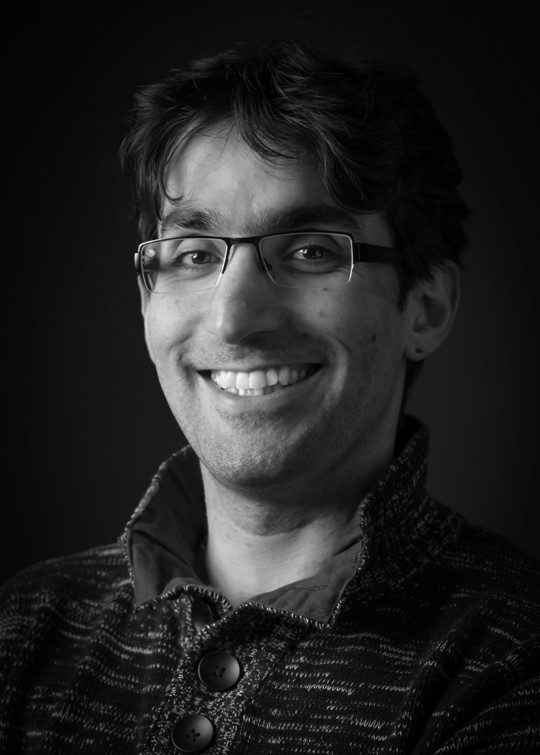}}]{Mehdi Darouich} 
	received the Ph.D. degree in embedded systems from the University of Rennes 1, Rennes, France, in 2010.
	He joined the IC Design and Embedded Software Division, CEA LIST (French Atomic Energy Commision), Saclay, France,
	where he works in the field of embedded processing architectures and real-time machine vision applications for embedded
	purposes. His current research interests include smart sensors architecture design, stereo vision perception, localization, and navigation.
\end{IEEEbiography}
\vskip -2\baselineskip plus -1fil
\begin{IEEEbiography}[{\includegraphics[width=1in,height=1.25in,clip,keepaspectratio]{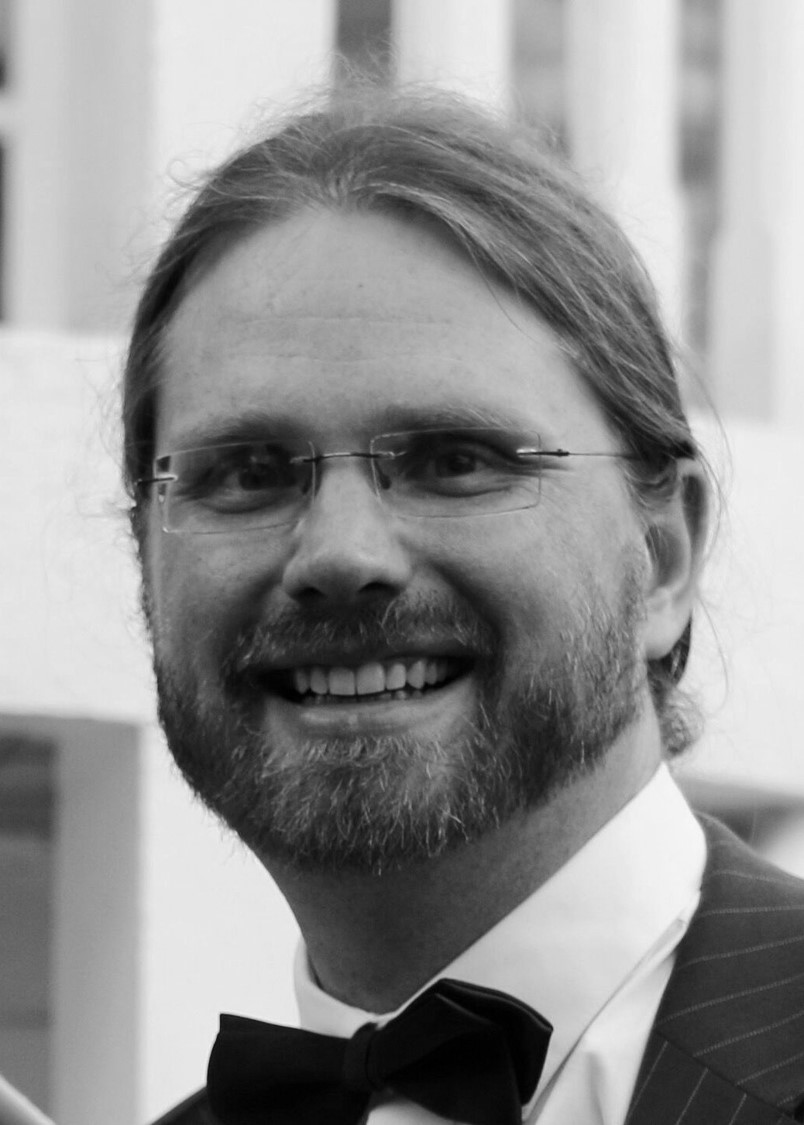}}]{Jean-Yves Didier} 
	is an Associate Professor at Universit\'e d'Evry-val d'Essonne (Universit\'e Paris-Saclay) since September
	2006. He received, in 2002, two M.S. degrees, both in computer science, from ENSIIE of Evry (public school of engineers)
	and Universit\'e d'Evry. He defended his Phd degree in robotics in 2005 at the same university. His research interests are
	focused on software architectures for mixed reality applications and their requirements such as localization and 3D environment
	reconstruction. He is also currently co-head of the Interaction, Virtual and Augmented Reality, Ambiant Robotics research team
	of the IBISC (Computer Science, Bio-Informatics and Complex Systems) laboratory of the University of Paris-Saclay, France.
\end{IEEEbiography}
\fi

\end{document}

%% file: abstract.tex
The 3D reconstruction of simultaneous localization and mapping (SLAM) is an important topic in the field for transport systems such as drones, service robots and mobile AR/VR devices. Compared to a point cloud representation, the 3D reconstruction based on meshes and voxels is particularly useful for high-level functions, like obstacle avoidance or interaction with the physical environment. This article reviews the implementation of a visual-based 3D scene reconstruction pipeline on resource-constrained hardware platforms. Real-time performances, memory management and low power consumption are critical for embedded systems. A conventional SLAM pipeline from sensors to 3D reconstruction is described, including the potential use of deep learning. The implementation of advanced functions with limited resources is detailed. Recent systems propose the embedded implementation of 3D reconstruction methods with different granularities. The trade-off between required accuracy and resource consumption for real-time localization and reconstruction is one of the open research questions identified and discussed in this paper.

%% file: introduction.tex
Autonomous transport systems, such as cars, service robots, UAVs/MAVs (unmanned/micro air vehicle) and mobile AR/VR (augmented reality/virtual reality) devices, require accurate and robust perception for high-level functions based on obstacle avoidance or interaction with its physical environment. Each transport system has several constraints with a different level of criticality for real-time processing. Autonomous cars offer much more space to include powerful and expensive computing hardware \cite{liu2020computing} than drones or AR/VR devices where the power consumption budget \cite{Chatzopoulos2017} and the robust localization \cite{COUTURIER2021103666} are critical. \par

Simultaneous Localization And Mapping (SLAM) is an active area of research and is widely used by the community to provide accurate and robust real-time localization and reconstruction of the surrounding environment without prior knowledge. It is reflected in three main questions \cite{Davison2018FutureMappingTC}: localization (where am I?), reconstruction (how is my environment?) and image segmentation (what are the objects around me?). The main challenge of SLAM is the global consistency. As it is mostly based on relative measurements from visual and inertial sensors, uncertainty accumulates gradually and the effect of drift begins to be noticeable over time. SLAM methods include an optimization module responsible for local and global consistency. The loop closure detection corrects the drift when a reconstructed scene has already been visited. Depending on the used techniques, SLAM offers several types of reconstruction, such as point clouds, surfels (oriented points), meshes (triangle meshes), volumetric (surface meshes based on voxels). This paper defines the term mapping as cartography for point clouds \cite{Campos2020, engel14eccv} and 3D reconstruction for mesh \cite{Rosinol20icra-Kimera} and volumetric models \cite{Oleynikova2016}. \par

While visual(-inertial) SLAM takes advantage of imaging sensors and inertial measurements, several lines of research make use of other sensors. In \cite{Alliez2020RealTimeMS, alliez:hal-02611679}, a multi-sensor system based on LiDAR \cite{Zhang2014} and a monocular infrared camera \cite{Kachurka2019} is used for real-time localization. In \cite{Vidal2018}, a hybrid state estimation pipeline combines event-based sensors, visible cameras and inertial measurements. Event-based sensors overcome the limitations of visual cameras against rapid movement or changes in lighting. Instead of capturing directly the light intensity, they acquire the change of intensity in the scene \cite{Gallego2020}. Today, they are mainly used as a complementary sensor to take advantage of the large amount of information provided by visible cameras. \par

Existing surveys on SLAM have reviewed the fundamental challenges for accurate and robust large-scale applications \cite{Cadena2016, Rosen2021, Stachniss2016}, from early probabilistic approaches and data association \cite{Durrant2006, Bailey2006} to the potential use of deep learning \cite{Chen2020}. SLAM components, including sensors to the embedded localization \cite{SALHI2019199} have been intensively studied to provide a robust solution to many applications, like autonomous driving \cite{Bresson2017}, search and rescue tasks, infrastructure inspection and 3D reconstruction in static and dynamic environments \cite{Zollhoefer2018} with challenging conditions \cite{Alkendi2021}. The robustness of real-time methods under difficult conditions has been reviewed and quantified including low visiblity \cite{Alkendi2021}, dynamic movement, illumination changes, changed viewpoints and lifelong scenarios \cite{bujanca2021robust}. Experiments \cite{bujanca2021robust} highlight that state-of-the-art approaches struggle with these challenging conditions. It also shows that the SLAM method based on feature extraction \cite{Campos2020} provide the best trade-off in terms of robustness. However, feature extraction for visual SLAM accentuates the lack of flexibility due to the dependence of a certain type of feature and the difficult localization in the presence of noise \cite{Azzam2020}. From the real-time localization to global mapping, the SLAM problem has been intensively studied using learning-based approaches \cite{Chen2020}. The evolution of deep neural networks (DNNs) and its impact on SLAM \cite{Cadena2016} opens several directions for lifelong scenarios including the type of model, scalability, and hardware deployment. \par

\noindent This survey provides a broader view of SLAM, from localization to 3D reconstruction in the embedded context, with an in-depth analysis of the implementation of advanced functions on resource-constrained hardware platforms. It focuses on methods using lightweight and low power consumption imaging sensors and inertial measurements. This article presents the following main contributions:

\begin{itemize}
    \item The description of each function for real-time localization and 3D reconstruction based on imaging sensors and inertial measurements in Section \ref{section:functions}. A discussion about strengths and limitations of existing visual(-inertial) SLAM methods as well as the potential use of deep learning is provided.
    \item A comprehensive review of the implementation of localization and reconstruction functions in low power consumption embedded systems with limited resources in Section \ref{section:hardware}.
\end{itemize}

%% file: figure_main.tex
\begin{figure*}[!t]
\centering
\includegraphics[width=\textwidth]{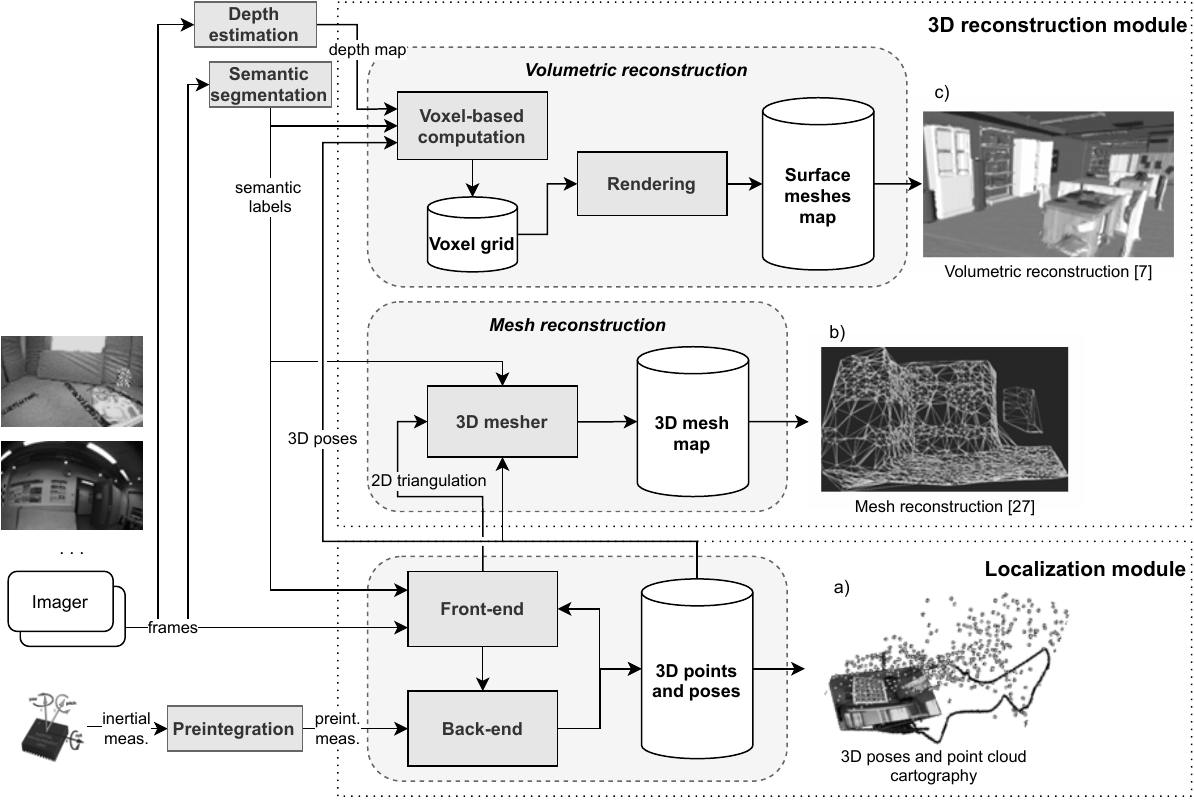}
\caption{Conventional real-time 3D scene reconstruction pipeline from imaging sensors and inertial measurements. a) represents an accurate trajectory estimation and a sparse point cloud scene representation. b) corresponds to the visualization of a 3D mesh introduced in \cite{Greene2017}. c) is the volumetric voxel-based reconstruction provided by \cite{Rosinol20icra-Kimera}.}
\label{fig:figure_main}
\end{figure*}

%% file: intro_s2.tex
The SLAM community has made remarkable improvement on the accuracy and robustness of large-scale applications in the recent years \cite{Rosen2021, Stachniss2016,Cadena2016}. \par

Figure \ref{fig:figure_main} illustrates a conventional 3D scene reconstruction pipeline using imaging sensors and inertial measurements as inputs. It is decomposed in two main modules, localization and 3D reconstruction. The first one is based on the front-end (FE) and the back-end (BE). The FE processes images and estimates motions. The BE uses preintegrated inertial measurements \cite{Forster2015} and manages the topological consistency through local and/or global optimizations of 3D poses and points \cite{Engels2006BundleAR}. The 3D reconstruction module provides a model of shapes and gives geometric properties. The mesh reconstruction allows a 3D mesh of the environment based on a 2D triangulation from FE and 3D poses provided by BE. This representation is useful for obstacle avoidance functions. The interaction with the physical environment requires a volumetric reconstruction provided by the voxel-based computation and the rendering of surface meshes. It takes as input the depth map from the image depth estimation and 3D poses from the BE. The semantic segmentation gives information about surrounding objets \cite{Garg2021, Kostavelis2015, Zhou2019}. For instance, in the case of a 2D RGB images, its semantic corresponds to a labelled classification of each pixel, used for volumetric reconstruction, mesh and point cloud cartography. \par

This section describes each part of the pipeline from real-time localization to 3D scene reconstruction and the potential use of deep learning. \par

%% file: intro_carto.tex
Real-time 3D reconstruction for mobile robots in an unknown environment requires an accurate localization that visual SLAM \cite{Campos2020, engel14eccv, Davison2003, Davison2007} and the subpart visual(-inertial) odometry methods (VO/VIO) \cite{Rosinol20icra-Kimera, Qin2019, Mourikis2007} provide from relative measurements.

%% file: frontend.tex
The front-end processes the input image. Indirect and direct techniques are the two main approaches, which are respectively described below. \par

Indirect methods consist of feature detection, feature matching (or tracking) and motion estimation from observations with geometric verification based on $n$-point random sample consensus (RANSAC) \cite{Nister2004, Horn87, Kneip2011}. Feature detection extracts corners on an $n$ gaussian pyramidal levels generated from the grayscale image. The process of the gaussian pyramid levels is downsizing an image through $n$ levels with the same factor ($640\times480$ to $80\times60$ in four levels with a factor of two) \cite{Adelson1984}. In \cite{Campos2020}, a scale factor of 1.2 accross 8 levels in the scale pyramid has been configured by default. This technique allows the feature extraction to be scale invariant. Several feature detection methods have been developed since Moravec \cite{Moravec1977} which give a trade-off between robustness and computing complexity \cite{harris1988combined, Shi1994, Rosten2006}. Feature matching involves detecting and describing features to match consecutive images \cite{Lowe2004, Bay2006, Rublee2011} whereas tracking with the KLT tracker estimates the displacement based on the detected features without any feature description \cite{Bouguet1999PyramidalIO}. The next step estimates the motion between two images and performs geometric verification with RANSAC for outlier rejection. Depending on the approach, features are specified in two or three dimensions for 2D-2D, 3D-3D and 3D-2D techniques. The first 2D to 2D case estimates the motion from the features extracted in 2D. This method is usually performed during the initialization using epipolar constraints to obtain geometric relations between two consecutive images \cite{Nister2004, Kneip2011, Civera2009}. In the case of 3D to 3D motion estimation, features are specified in three dimensions based on the triangulation of 3D points by using, for instance, stereo visible cameras. The 3D-3D estimation is based on minimizing the Euclidean distance between corresponding 3D features. Finally, the 3D to 2D motion estimation approach is based on the perspective from $n$ points (PnP) which minimizes the reprojection error between the 2D feature and its 3D counterpart:
\begin{equation}
    \label{eqn:reprojerror}
    \MakeUppercase{\textbf{T}}_{k} = arg\underset{T}{min}\sum_{i} \lVert u'_{i} - \pi(p_{i}) \rVert^2
\end{equation}
where $\MakeUppercase{\textbf{T}}_{k}$ is the transformation matrix between the current view point and an arbitrary origin, $u’_{i}$ the detected features, $\pi$ the projection function and $p_{i}$ the corresponding 3D point. As pointed in \cite{Nistera}, the 3D to 2D motion estimation is more accurate than 3D-3D estimation because of the uncertainty generated by the triangulated 3D points.\par

While indirect methods extract and use features for the pose estimation, direct methods use all pixels available in the input image \cite{engel14eccv, Newcombe2011a, Engel2016, Cremers2017}. The pipeline differs as they do not contain the feature detection and matching stages. It estimates an optical flow with the photometric error minimization:
\begin{equation}
    \label{eqn:photoerror}
    \MakeUppercase{\textbf{T}}_{k,k-1} = arg\underset{T}{min}\sum_{i} \lVert I_{k}(u'_{i}) - I_{k-1}(u_{i}) \rVert^2 
\end{equation}
where $\MakeUppercase{\textbf{T}}_{k,k-1}$ is the transformation matrix between two frames, $I_{k}(u'_{i})$ the intensity $I$ of the pixel $u'_{i}$ in frame $k$.\par

Table \ref{tab:table_techniques} summarizes the benefits and drawbacks of indirect and direct approaches. Indirect methods allow robust and efficient computation due to a sparse detection of features. However, they rely on detected features and require robust estimation to prune outliers that impair the localization process. Direct methods have the benefits to estimate motion by minimizing a photometric error. However, they require an accurate photometric calibration. In \cite{Engel2016}, it takes into account exposure time, a non-linear response function and lens vignetting which allows a comprehensive model of the brightness transformation. In the presence of a large amount of intensity gradients from the image, low computational speed is one of the drawbacks for direct methods. Dense representations with direct image alignement used the commodity of GPU hardware to achieve real-time performances \cite{Newcombe2011a, Pizzoli2014}. In order to reduce the computation time, a semi-dense depth filtering \cite{Engel2013} has been developed on CPUs \cite{engel14eccv, engel2015_stereo_lsdslam}. In \cite{Engel2016}, a direct and sparse model uses only a selection of well-distributed points in the image with a high image gradient magnitude. \par

In order to combine the strengths of direct and indirect methods, semi-direct approaches exploit pixels with strong gradients while relying on sparse features \cite{Dong2021, forster2016svo}. \par

Real-time processing for these approaches is achieved through the use of keyframe (KF)-based techniques to provide a balance between accuracy and efficiency \cite{klein07parallel}. The front-end module estimates if an input frame is considered as a keyframe for back-end optimization. Poses and 3D points are produced over time by these keyframes to obtain a local and/or global consistency. Keyframes are generated at a lower framerate (8-16 FPS) than input frames (30-60 FPS).

%% file: table_techniques.tex
\begin{table*}[!ht]
\renewcommand{\arraystretch}{1.3}
\caption{Description, benefits and drawbacks of both indirect and direct approaches used in state of the art methods.}
\label{tab:table_techniques}
\centering
%\resizebox{\textwidth}{!}{
\begin{tabular}{|l|l|l|}
\hline
\multicolumn{1}{|c|}{} & \multicolumn{1}{c|}{\textbf{Indirect approach}}                                                                                                                                       & \multicolumn{1}{c|}{\textbf{Direct approach}}                                                                                                               \\ \hline \hline
Description            & \begin{tabular}[c]{@{}l@{}}- Feature extraction\\ - Feature matching or tracking\\ - Reprojection error minimization\end{tabular} & \begin{tabular}[c]{@{}l@{}}- Uses all pixels\\ - Photometric error minimization\end{tabular} \\ \hline
Benefits             & \begin{tabular}[c]{@{}l@{}}- Robust and fast due to feature extraction\\ - Allows sparse reconstruction with efficient computation\end{tabular}                                                       & \begin{tabular}[c]{@{}l@{}}- More robust and accurate than indirect\\ - Allows dense scene representation\end{tabular}           \\ \hline
Drawbacks         & \begin{tabular}[c]{@{}l@{}}- Dependant to features for matching\\ - Requires robust estimation to overcome outliers\end{tabular}                                                                      & \begin{tabular}[c]{@{}l@{}}- Longer computation time due to optical flow estimation\\ - Requires a good initialization\\ - Requires an accurate photometric calibration\end{tabular} \\ \hline
\end{tabular}
%}
\end{table*}

%% file: backend.tex
In early work, SLAM methods were composed of, an extended Kalman filter (EKF) \cite{Leonard1990, smith1990estimating, Leonard1991} and with a separate back-end for bundle adjustment \cite{klein07parallel} and in the form of a node graph \cite{Folkesson2004, Frese2006, Dellaert2006}. The optimization back-end function is a crucial step to obtain an accurate pose estimation related to constraints of new measurements. It consists of optimizing the map composed of 3D poses and points in a global or local manner. This subsection addresses the challenge of optimization as follows: First, the input of inertial measurements in the BE. Second, the approaches used to optimize states and 3D points of the map. Then, the functions to provide global map consistency, which is identified as a fundamental key topic in SLAM \cite{Cadena2016, Aulinas2008}.\par

In a visual-inertial system, inertial measurements are preintegrated to improve localization by solving the scale factor problem through linear acceleration and angular velocity between frames. IMUs provide data at high rate (100 Hz to 1 kHz) compared to the framerate (20-60 FPS). To cope with the acquisition difference, a preintegration method is included in a real-time localization pipeline that analyses and consolidates all inertial data between two keyframes into a single measurement \cite{Forster2015}. This technique uses a structureless approach which avoids optimizing over the 3D points to improve the computation time during optimization. \par

In order to fuse visual and inertial data and optimize states generated by the localization module, two main approaches are used, filtering-based or factor graph \cite{Gui2015}. In a filtering system, only the latest state is estimated. The EKF fuses data provided by different sensors, predicts the future state with respect to the initial estimation and updates the prediction. The uncertainty is represented using a covariance matrix. The complexity of this approach increases over time as the generated map and estimated 3D points become larger. Therefore, the memory consumption increases significantly in addition to the computational cost. In \cite{Davison2007}, real-time processing has been reached by taking into account only a small number of features. In \cite{Bloesch2015}, direct photometric errors have been used within the EKF update and employed a numerical minimal distance representation of features to address the computational issue. Alternative methods based on the Multi-State Constraint Kalman Filter (MSCKF) framework \cite{Mourikis2007} use a structureless strategy that marginalizes the 3D points \cite{Sun2018, Geneva2020}. \par

Another optimization technique is the factor graph \cite{Dellaert2017}, which represents all states, points, data related to each other in the form of a nonlinear graph \cite{gtsam, Kaess2011, Kummerle2011, ceres-solver}. This graph is solved to optimize local (\textit{fixed-lag smoothing}) and/or global (\textit{full smoothing}) states and 3D points while exploiting the sparsity of SLAM algorithms. In a fixed-lag smoother configuration, only the poses within a sliding time window are optimized while a full smoother takes into account the full history of poses. The optimization based on the factor graph is described as follows:
\begin{enumerate}
    \item Linearization of the factors in the graph (IMU, visual data, etc.) in a linear equation system 
    \begin{equation} 
        H\Delta x=\varepsilon 
    \end{equation} 
    where $H$ is the Hessian matrix, $\varepsilon$ is a vector that describes how front-end measurements affect the state of each keyframes and $\Delta x$, a vector describing the updated states. 
    \item Use of the Cholesky factorization and back-substitution to solve the linear equation system. 
    \item Marginalization of states outside the sliding temporal window for local optimization to maintain real-time performances.
    \item Solutions of the linear system are used to update the remaining states of keyframes.
\end{enumerate} \par
In \cite{Usenko2020}, a set of non-linear visual-inertial information about the motion estimation between keyframes are recovered from the first layer odometry and combined using a global bundle adjustment. Based on a nonlinear optimization method \cite{Qin2018}, VINS-Fusion \cite{Qin2019} supports the use of multiple sensors (camera, IMU, GPS, etc.) integrated in a pose graph structure as a Maximum Likelihood Estimation (MLE) problem. In \cite{Leutenegger2013}, the nonlinear optimization integrates both the reprojection errors and a temporal error term from inertial measurements while old keyframes are marginalized from the optimization window to ensure real-time processing. The odometry module of Kimera-VIO \cite{Rosinol20icra-Kimera} is based on the IMU preintegration approach \cite{Forster2015} and provides SLAM capabilities with a pose graph optimization module responsible for loop closures. \par

Global consistency in SLAM is provided with loop closures to detect that a scene has already been visited and to correct the accumulated drift. Without it, visual SLAM becomes visual odometry for local consistency \cite{Cadena2016}. Loop closure is performed in three main steps: 
\begin{enumerate}
    \item Detect candidates between newly created features from poses and the current active map
    \item Correct the detected poses affected by the loop closing
    \item Optimize the map in order to verify if the accumulated drift has been corrected
\end{enumerate}
The mainstream method for loop closure is the use of a bag-of-words (BoW) vector. It implements a recognition database of visual words vocabulary describing image features \cite{Galvez-Lopez2012}. Therefore, the recognition database is queried to find loop candidates. In terms of timing, loop closures of the visual-inertial ORB-SLAM3 system \cite{Campos2020} takes around 10ms for the detection, 124.77ms for the correction, which includes the loop closure and the correction of the whole map and 1529.69ms for the full map optimization. In order to maintain real-time performances, the latter is only performed if the number of keyframes to be optimized is below a fixed threshold. A common conclusion is that the extraction and comparison of feature descriptors require too much computational resources. Therefore, alternatives methods have been developed, such as the extraction of the shape of each object as binary content \cite{Wang2019} and edge-based verification for loop closures \cite{Schenk2019}. The latter provides an average accuracy of 2.36 cm, which is slightly worse compared to 1.14 cm with ORB-SLAM2 \cite{Mur-Artal2017} on RGB-D dataset \cite{Sturm2012}.

%% file: intro_recons.tex
While the localization module provides a point cloud cartography, the 3D reconstruction module considers two types of map representation, the mesh and the volumetric reconstruction based on voxels to produce surface meshes in real-time.

%% file: mesh.tex
Mesh representation is widely used to model surfaces, shapes and provides a topology of the scene based on points \cite{Rosinol20icra-Kimera} or surfels \cite{Schps2020SurfelMeshingOS}. The Delaunay triangulation has been used for many applications, notably in computer vision \cite{Dinas2014} to provide accurate mesh reconstruction and cover potentially usable planar surfaces \cite{Rosinol2019}. Several algorithms derive the Delaunay triangulation in two \cite{Rosinol20icra-Kimera, Greene2017, Teixeira2016, Yokozuka2019VITAMINEVT} and three \cite{Piazza2018} dimensions. In 2D, this technique follows the empty circle property, which generate a triangle when the circle is the only one to pass through all three vertices. Therefore, there are no vertices in it. The triangulation maximizes the minimum angle of each generated triangles. Thus, the reconstruction ensures the consistency of triangulations and high quality meshes. \par

In \cite{Greene2017}, a lightweight Delaunay mesh \cite{Shewchuk2002} is generated with a keyframe-less approach and monocular depth estimation for MAVs. The reconstruction provides a per-frame mesh reconstruction without taking into account the previous meshes for fast computation. The approach was able to process each frame onboard an MAV with an Intel Skull Canyon NUC flight computer at over 90 Hz. In \cite{Rosinol20icra-Kimera}, the multi-frame mesh gives a single mesh built over time on the basis of fusing per-frame meshes. On an Intel Xeon CPU E3-1505M v6 3 GHz, the multi-frame mesh is generated at around 67 Hz. This configuration shows that a reliable map composed of a large mesh built over time is maintained and updated with each new mesh generated in real-time.

%% file: volum.tex
A volumetric reconstruction is defined as a voxel-based computation in which a surface mesh is rendered from the volume to allow a detailed 3D model \cite{Lorensen1987}. 

\subsubsection*{Voxel-based computation}
A common method to quickly and accurately represent surfaces is the truncated signed distance field (TSDF) \cite{CurlessL96}. This technique represents the 3D environment as a grid of voxels. In a TSDF volume, which integrates depth data, the value of each voxel corresponds to the signed distance to the nearest surface. Positive and negative values correspond respectively to voxels outside or inside the volume. Thus, the surface is defined by the isosurface boundary between negative and positive values (zero-crossing). Beyond a certain distance, the information becomes irrelevant. Therefore, the distance is truncated to take advantage of the values close to the surface. \par

Each voxel stores a truncated signed distance $D$ and weight values $W$ updated for each 3D point $p$ in the volume from frames $1...k$.

\begin{equation}
	D(p) = \frac{\sum w_{k}(p)d_{k}(p)}{\sum w_{k}(p)}
\end{equation}

\begin{equation}
	W(p) = \sum w_{k}(p)
\end{equation}
where $d$ is the signed distance and $w$ the weight function of the sensor measurements. The cumulative $D_{k}(p)$ and $W_{k}(p)$ are expressed as follows \cite{CurlessL96}:

\begin{equation}
    D_{k+1}(p) = \frac{W_{k}(p)D_{k}(p)+w_{k+1}(p)d_{k+1}(p)}{W_{k}(p)+w_{k+1}(p)}
\end{equation}

\begin{equation}
    W_{k+1}(p) = W_{k}(p) + w_{k+1}(p)
\end{equation}

The choice of the weighting have a strong impact on the accuracy of the representation as it models the uncertainty of surface measurements \cite{Oleynikova2016}. Therefore, the greater the uncertainty, such as noisy data, the higher the weighting. For instance, to allow real-time 3D reconstruction, KinectFusion \cite{Newcombe2011} adopted a common constant weight approach $w_{k}(p) = 1$ resulting in a simple average. 

\begin{equation}
    W_{k+1}(p) = min(W_{k}(p) + w_{k+1}(p), W_{max})
\end{equation}

\noindent In \cite{Oleynikova2016}, a comparison of two weighting strategies is proposed. It consists in comparing the 3D reconstruction qualitatively and quantitatively with a constant and a quadratic weight. Equation \ref{eq:weightvoxblox} defines the truncated distance so that $\delta = 4v$ and $\epsilon = v$, where $v$ is the voxel size.

\begin{equation}
	\label{eq:weightvoxblox}
	w(p) =
	\begin{cases}
		\frac{1}{z^2},& -\epsilon < d \\
		\frac{1}{z^2}\frac{1}{\delta-\epsilon}(d + \delta),& -\delta < d < -\epsilon \\
		0,& d < -\delta
	\end{cases}
\end{equation}

\noindent Qualitatively, the quadratic weighting strategy provides a better representation of the structure with less error than constant weights. Higher level of robustness with less distorsion can be observed with a voxel size $v$ ranging from 0.02m to 0.2m.

\subsubsection*{Rendering}

In order to render the voxel 3D perspective, two main approaches are used for mesh extraction, raycasting and projection mapping \cite{Klingensmith2015}. Raycasting casts a ray from sensor to the truncated signed distance behind the 3D point within the voxel and extracts the zero-crossing from the TSDF values to obtain an approximate depth rendering \cite{CurlessL96, Parker1998}. Projection mapping calculates the distance between the center of the voxel and the depth value in the image. The amount of data processed by the projection mapping technique is bounded by the number of voxels while the raycasting one is potentially unbounded as it includes the number and the length of rays being integrated. Timing results shows that raycasting is faster than projection mapping with a required time of about 62ms and 106ms respectively on a Tango Yellowstone tablet device with quadcore CPU, 4GB RAM and a Tegra K1 GPU \cite{Klingensmith2015}. \par

In pioneer work \cite{Newcombe2011}, the Kinect sensor has been used to provide RGB frames and depth measurements at a $640\times480$ resolution. It uses the TSDF integration and 3D rendering with raycasting. The 6DoF pose is computed with the iterative closest point (ICP) approach \cite{Besl1992}. The volumetric reconstruction is limited in area of a $512^3$ volume. The implementation on NVIDIA GeForce GTX 560 updates TSDF values at 2ms \cite{izadi2011kinectfusion} and the main system operates in real-time at 40 FPS with a resolution of $512^3$. Fully implemented with CUDA in order to use NVIDIA GPUs, this method is not memory efficient with 512 MB allocated for 32-bit voxels in the restricted space. A voxel hashing method \cite{Niesner2013} has been used in \cite{Kahler2015} to obtain a memory efficient and large scene representation. The algorithmic approach introduced intermediate online visualization to speed up the calculation. While raycasting is performed for each frame in a full configuration, a forward projection parameter uses the most recent raycasting result. After a fixed threshold of $n$ frames a full raycast is performed. Implemented on NVIDIA Tegra K1 and Apple iPad Air 2, the method achieves a computation time per frame of 21.04ms and 48.43ms respectively. \par

In \cite{Oleynikova2016}, the same hierarchical data structure with voxel hashing \cite{Niesner2013} as well as a grouped raycasting approach has been used. The latter merges the cast point with all other 3D points to the same voxel to perform raycasting only once, on the average position. It needs around 55ms and 5ms to process a voxel size of 0.05m and 0.20m respectively compared to 250ms and 100ms with a standard approach.

%% file: intro_dnn.tex
Advanced research in deep neural networks (DNN) allows the use of a large set of data while providing accurate and robust results for many applications \cite{sze2017efficient}. This subsection provides an overview of DNN methods for real-time localization, a comprehensive description of intermediate representations and the complexity of deep learning in the context of embedded systems.

%% file: dnn_slam.tex
Several research papers discussed the use of neural networks for real-time pose estimation with an end-to-end or hybrid implementation. The first one only relies on deep learning to estimate the pose from consecutive images. In \cite{Wang2017}, a convolutional neural network (CNN) \cite{Dosovitskiy2015} was used to extract features from RGB images, and a recurrent neural network (RNN) to model sequential information. Therefore, long short-term memory (LSTM) is used for the sequential representation with a retained memory of previous hidden states. In the case of a visual-inertial configuration, a multi-rate LSTM has been implemented to process inertial measurements \cite{Clark2017VINetVO}. End-to-end methods based on supervised \cite{Wang2017} or unsupervised \cite{Li2021} learning are evolving, but the performances of current approaches are limited by training data to provide an accurate and robust estimation in all situations. \par

A hybrid pipeline uses deep learning for specific back-end (BE) and front-end (FE) functions, including local \cite{tang2018banet, Czarnowski2020DeepFactorsRP} or global \cite{Arshad2021, Zhang2017CNN} BE optimization and FE feature extraction \cite{Li2020, Sons2019, Xu2020, zhou2021patch2pix}. In \cite{Tang2019}, results based on a CNN detector and descriptor demonstrated better distributed features with a lower number of detection compared with ORB \cite{Rublee2011}. The extraction runs at 40 FPS while the SLAM pipeline based on ORB-SLAM2 \cite{Mur-Artal2017} runs at 20 FPS on Jetson TX2 with a tiny version of the network. The accuracy of the hybrid method deteriorates for sequences that require fine details, like the \textit{fr1\_floor} and \textit{fr1\_360} of the TUM RGB-D dataset \cite{Sturm2012} due to the scale of feature maps.

%% file: semantic.tex
Intermediate representations have been proven useful for high-level functions based on pixels-to-action models. In a visual-based urban driving task where an agent has to reach a target location, success accuracy improved by around 15\% with the addition of depth estimation and by 20\% with semantic segmentation compared to an RGB image only \cite{Zhou2019}.

\subsubsection*{Semantic segmentation}
In order to recognize surrounding objects, the image segmentation, like semantics, allows a set of objects to be labelled at pixel-level \cite{Garg2021, Minaee2021}. \par

In \cite{Xia2020}, the concept of semantics for object recognition and semantic segmentation in SLAM is detailed and argues that it provides effective solutions for data association \cite{Lianos2018} and long-term consistency \cite{Gawel2018} to obtain a reliable localization. Deep learning techniques enhance the capabilities of SLAM by using semantic segmentation for local loop closure detection and medium-term consistency \cite{Lianos2018, Zhang2020} or for handling dynamic environments. As conventional static SLAM considers dynamic features as outliers, semantics is used to label dynamic and static elements to improve the accuracy and robustness of localization \cite{Wen2021, Yu2018}. Semantic extraction has also known active research for the segmentation of the map, which leads to real-time object recognition based on point clouds \cite{Tateno2015}, surfels \cite{McCormac2017, Wald2018}, meshes \cite{Rosinol20icra-Kimera, Rosu2019} and voxels \cite{Rosinol20icra-Kimera, Grinvald2019, Narita2019} representations. Semantic reconstruction is achieved by two main approaches, view-based or map-based labelling \cite{Landgraf2020}. Semantic extraction per view processes raw 2D input image data but performs unnecessary computations from one view to another whereas the segmentation of the whole map involves a 3D CNN. It avoids redundant computation but depends on the quality of the reconstruction. Using the mean intersection over union (IoU) metric, the segmentation achieves an accuracy of 89\% and 92\% after 1000 frames for view-based and map-based respectively \cite{Landgraf2020}. \par

Several questions remain open, including the accuracy of the segmentation required for a 3D reconstruction pipeline and its impact on the precision of the reconstruction. The complexity of semantic generation in terms of computational cost and use of hardware resources for 2D or 3D segmentation tasks is also part of future open directions in the embedded context.

\subsubsection*{Depth estimation}
Deep neural networks have been used for depth estimation of monocular images \cite{Czarnowski2020DeepFactorsRP, Tateno2017, Bloesch2018}. Depth estimation based on a CNN overcomes the limitations of monocular cameras by handling pure rotational motions \cite{Tateno2017} and resolving the scale factor problem by updating the transformation matrix with the scale parameter $\alpha$ \cite{Yin2017}:

\begin{equation}
    \label{eq:scaleproblem}
    \MakeUppercase{\textbf{T}}_{k,k-1} = 
        \begin{bmatrix}
            \MakeUppercase{\textbf{R}}_{k,k-1} & \alpha\textbf{t}_{k,k-1} \\
            0 & 1
        \end{bmatrix}
\end{equation}
where $\MakeUppercase{\textbf{T}}_{k,k-1}$ is the transformation matrix between two frames (rotation $\MakeUppercase{\textbf{R}}_{k,k-1}$, translation $\textbf{t}_{k,k-1}$) and $\alpha$ the scale parameter defined as a maximum likelihood estimation taking into account the estimated depth value of each pixel. \par

Although this representation has several benefits in the case of monocular SLAM, its accuracy remains limited for long sequences, compared to traditional stereo methods with a strongly drifting trajectory \cite{Li2018}. In \cite{Martins2018}, the depth map has been estimated from the fusion of a dense stereo pipeline with a monocular depth estimation based on a self-supervised CNN with data generated from the sensors. Qualitatively, the fused map refines the noisy result of the stereo vision and gives the global composition of the image provided by the monocular estimation. The median absolute error between depth estimation and the ground truth (GT) depth distance for the stereo and mono is less than 5 m and around 20 m respectively for a GT of 80 m and around 1 m and less than 4 m respectively for a GT of 10 m. \par

In the embedded context with restrictive constraints, the right trade-off between required depth accuracy for 3D reconstruction and computational cost in terms of memory usage and processing time remains the main open line of research.

%% file: complexity_dnn.tex
The complexity of deep learning models for real-time localization and reconstruction can be characterized by three main points: the type of model, the use of large amounts of data for scalability, the resource consumption and hardware implementation for real-time processing. End-to-end models are currently computationally expensive, as they involve the sequential representation of previous states. Hybrid models have the advantage of combining the strengths of deep learning with the maturity of conventional methods for specific functions to overcome several sensors limitations and to compute intermediate representations for high-level understanding of the surrounding environment.\par

The accuracy of deep learning methods is highly dependent on the training data. However, in order to obtain a reliable system, the training stage must learn from many situations, especially for real-time localization. Visual odometry methods based on DNNs can fail, resulting in low accuracy if the model is not well-fitted \cite{Wang2017}. The majority of the localization methods are trained and evaluated on the KITTI benchmark \cite{Geiger2012CVPR}, which provides car driving trajectories without significant rotation changes. This can lead to significant errors in different situations. \par

In addition to the time and resources required for the training stage of deep neural networks, specific hardware platforms allow real-time processing for DNN inference, especially with the parallelization of computations for better use of GPUs. Hybrid pipelines take advantage of both the CPU and the GPU for geometric constraints and neural network calculations, respectively \cite{Czarnowski2020DeepFactorsRP}. For real-world deployment, size, power consumption and resource constraints must be considered. In \cite{Yu2020}, feature extraction with a CNN has been accelerated on FPGA in fixed-point number \cite{Yu2018FPGA} and the visual odometry network \cite{Zhan_2018_CVPR} is run on CPU. This implementation achieves an execution time of 5ms on FPGA and 340ms on CPU with a monocular camera at 20 FPS.

%% file: intro_method.tex
This subsection provides a description of the benchmarking tools for qualitative and quantitative evaluation of state-of-the-art methods. It also highlights the techniques used, from real-time localization to DNNs, in existing approaches.

%% file: benchmark.tex
Many benchmarking tools present various sequences recorded at different locations for evaluation of visual-based methods under challenging conditions such as illumination changes, long trajectories, different motion speeds and dynamic environments. \par

Depending on the transport use cases, public datasets provide outdoor environments for autonomous cars with densely populated area \cite{Wen2020UrbanLocoAF, Hsu2021}, diverse urban environments \cite{jjeong-2019-ijrr} and large scale sequences with different speeds up to 80 km/h \cite{Geiger2012CVPR}. Dynamic and long-term environments with illumination changes \cite{shi2019openlorisscene, Carlevaris2016} are available for service robots, where human activity is also included \cite{pronobis2009ijrr}. Challenging motions indoors and outdoors are provided with MAV's datasets through aggressive trajectories \cite{Delmerico19icra}, fast motion and motion blur \cite{Burri2016} and in urban streets at low altitude (5-15 m above the ground) \cite{Majdik2017}. \par

Data recorded with a handheld sensor is particularly useful for AR/VR applications in dynamic sequences \cite{palazzolo2019iros} and global/local illumination changes \cite{park2017icra}. The TUM datasets provide several handheld trajectories with many imaging sensors for short and long sequences \cite{Schubert2018, Engel2016a, sturm12iros}. Capturing data in real-world scenes has several limitations \cite{Wang2020} that can be overcome with synthetic sequences. The accuracy of surface reconstruction with RGB-D information under realistic \cite{handa:etal:ICRA2014} and challenging conditions \cite{park2017icra, Rosinol21arxiv-Kimera} is also evaluated on the basis of synthetic datasets. \par

The evaluation of the estimated trajectory and the accuracy of the scene reconstruction requires ground truth usually provided by an RTK-GPS or motion capture systems.

%% file: table_slam.tex
% \usepackage{multirow}
%\begin{table*}[!t]
\begin{sidewaystable*}[]
    \renewcommand{\arraystretch}{1.2}
    \caption{Existing visual(-inertial) odometry, SLAM and 3D reconstruction methods with the potential use of deep neural network (DNN) and the hardware (HW) implementation.}
    \label{tab:table_slam}
    \centering
    \resizebox{\textwidth}{!}{
    \begin{tabular}{lccccccccccll}
    \hline
    \multicolumn{1}{c}{\multirow{3}{*}{\textbf{Methods}}} & \multirow{3}{*}{\textbf{Year}} & \multicolumn{4}{c}{\textbf{Sensors}}                                                                            & \multicolumn{2}{c}{\multirow{2}{*}{\textbf{Localization strategy}}} & \multicolumn{3}{c}{\multirow{2}{*}{\textbf{Type of representation}}} & \multicolumn{1}{c}{\multirow{3}{*}{\textbf{Usage of DNN}}} & \multicolumn{1}{c}{\multirow{3}{*}{\textbf{HW implementation}}} \\ \cline{3-6}
    \multicolumn{1}{c}{}                                  &                                & \multicolumn{2}{c}{\textbf{Visible cam.}} & \multirow{2}{*}{\textbf{IMU}} & \multirow{2}{*}{\textbf{Others}} & \multicolumn{2}{c}{}                                                & \multicolumn{3}{c}{}                                                 & \multicolumn{1}{c}{}                                        & \multicolumn{1}{c}{}                                   \\ \cline{3-4} \cline{7-11}
    \multicolumn{1}{c}{}                                  &                                & \textbf{mono}        & \textbf{stereo}        &                               &                                  & \textbf{Front-end}               & \textbf{Back-end}                 & \textbf{Carto.}        & \textbf{Mesh}   & \textbf{Volum.}   & \multicolumn{1}{c}{}                                       & \multicolumn{1}{c}{}                                   \\ \hline
    DeepSLAM \cite{Li2021} & 2021 & \cmark & \cmark & - & - & \multicolumn{2}{c}{End-to-end DL model} & Dense & - & - & Depth (48ms), poses (25ms), loop (120ms) & i7-6820HK 2.7GHz, GeForce GTX 980 M (20 FPS)\\ \hline
    ORB-SLAM3 \cite{Campos2020}                                               & 2020                           & \cmark                    & \cmark                      & \cmark                             & -                                  & Det.+Des.                      & Graph                             & Sparse    & -                 & -                       & -                                        & \begin{tabular}[c]{@{}l@{}}\cite{Mur-Artal2017} on Jetson TX2 (28 FPS) \cite{Aldegheri2019}\\ \cite{Mur-Artal2017} on Pi 3B+ (6 FPS), Jetson Nano (10 FPS) \cite{Silveira2020}\end{tabular}                                                         \\ \hline
    Basalt \cite{Usenko2020}                                  & 2020                           & -                      & \cmark                      & \cmark                             & -                                  & Tracking                         & Graph                             & Sparse                             & -                 & -                       & -                                        & E5-1620 CPU (19 FPS global opti., 128 FPS local opti.)                                                         \\ \hline
    OpenVINS \cite{Geneva2020}                                                & 2020                           & \cmark                      & \cmark                        & \cmark                               & -                                  & Tracking                         & Filtering                         & Sparse                             & -                 & -                       & Depth estimation \cite{Merrill2021ICRA}                                        & Jetson TX2 (36 FPS), Jetson Nano (28 FPS) \cite{Merrill2021ICRA}                                                         \\ \hline    
    DXSLAM \cite{Li2020}    & 2020  & \cmark    & - & - & - & Det.+Des.   & Graph  & Sparse    & - & - & Feature extraction (46.2ms w/ opti.) & Core i7-10710U (15W)   \\ \hline
    \cite{Alliez2020RealTimeMS}                                         & 2020                           & \cmark                      & -                        & \cmark                             & IR+LiDAR                         & Det.+Des.                      & Graph                         & Sparse/Dense              & -                 & Offline \cite{Verdie2015}                       & -                           & NI IC-3173 (10 FPS visual-SLAM, 20 FPS lidar-SLAM)                                                         \\ \hline    
    SurfelMeshing \cite{Schps2020SurfelMeshingOS}                                           & 2020                           & -                    & -                        & -                               & RGB-D                            & -                           & -                                 & Surfels                             & \cmark               & -                       & -                                        & Core i7 6700K,  Geforce GTX 1080 (178 FPS for mesh)                                                         \\ \hline
    Kimera \cite{Rosinol20icra-Kimera}                                                  & 2020                           & \cmark                      & \cmark                      & \cmark                             & -                                  & Tracking                         & Graph                             & Sparse                      & \cmark               & \cmark                     & Semantic segmentation                               & Jetson TX2 \cite{Rosinol21arxiv-Kimera}                                                         \\ \hline    
    DeepFactor \cite{Czarnowski2020DeepFactorsRP}   & 2020  & \cmark    & - & - & - & Semi-direct   & Graph & - & - & \cmark & Depth estimation    & \begin{tabular}[c]{@{}l@{}}GTX 1080 GPU (network, camera tracking)\\CPU (geometric error factors)\end{tabular} \\ \hline        
    \cite{Xu2020}   & 2020  & \cmark    & \cmark    & - & - & Det.+Des.   & - & - & - & - & Feature extraction (59ms) & Xilinx ZCU102 FPGA    \\  \hline    
    ST-VIO \cite{Zhang2020} & 2020  & \cmark    & - & \cmark & - & Direct & Graph & - & - & - & Semantic segmentation \cite{Wang2019Segmentation} & Jetson TX2 \\ \hline
    \cite{Yu2020}   & 2020 & \cmark & - & - & - & \multicolumn{2}{c}{End-to-end DL model} & - & - & - & \begin{tabular}[c]{@{}l@{}}Feature extraction on FPGA (5ms)\\VO on ARM CPU (340ms)\end{tabular} & Xilinx ZCU102 MPSoC \\ \hline        
    GCN-SLAM \cite{Tang2019} & 2019 & - & - & - & RGB-D & Det.+Des. & Graph & Sparse & - & - & Feature extraction (25ms) & Jetson TX2 (20 FPS) \\ \hline
    VITAMI-E \cite{Yokozuka2019VITAMINEVT} & 2019 & \cmark & - & - & - & Tracking & Graph & Dense & \cmark & \cmark & - & Core i7-7820 HQ CPU \\ \hline
    RESLAM \cite{Schenk2019} & 2019 & - & - & - & RGB-D & Edge & Graph & Sparse & - & - & - & i7-4790 desktop computer with 32 GB RAM \\ \hline
    \cite{Sons2019} & 2019  & \cmark    & \cmark    & - & - & Det.+Des.   & Graph  & Sparse    & - & - & Feature extraction (around 16ms)    &  GeForce GTX Titan X  \\ \hline    
    VINS-Fusion \cite{Qin2019}                                             & 2019                           & \cmark                    & \cmark                      & \cmark                             & LiDAR                            & Tracking                         & Graph                             & Sparse                             & -                 & -                       & -                                        & \cite{Qin2018}: UP Board (7 FPS), ODROID XU4 (7 FPS) \cite{Delmerico2018}                                                         \\ \hline 
    PanopticFusion \cite{Narita2019}                                          & 2019                           & -                    & -                      & -                             & RGB-D                                & -                                & -                                 & -                             & -                 & \cmark                     & Semantic segmentation (315ms) \cite{He2017}                    & Image segmentation on GPU                         \\ \hline    
    MSCKF-based \cite{Sun2018}                                        & 2018                           & -                      & \cmark                        & \cmark                               & -                                  & Tracking                         & Filtering                         & Sparse                             & -                 & -                       & -                                        & \cite{Zhu2017}: UP Board (20 FPS), ODROID XU4 (20 FPS) \cite{Delmerico2018}                                                          \\ \hline 
    \cite{Piazza2018}                                         & 2018                           & -                    & -                      & -                             & -                                & -                                & -                                 & -                             & \cmark               & -                       & -                                        & Core i7-4770S, 3.10 GHz with \cite{Mur-Artal2015}                                                         \\ \hline    
    DS-SLAM \cite{Yu2018} & 2018 & - & - & - & RGB-D & Det.+Des. & Graph & - & - & \cmark & Semantic segmentation (37.6ms) \cite{Badrinarayanan2017} &  Intel i7 CPU, P4000 GPU \\ \hline        
    CNN-SLAM \cite{Tateno2017}                                                & 2017                           & \cmark                    & -                        & -                               & -                                  & Direct                           & Graph                             & Dense      & -                 & -                       & \begin{tabular}[c]{@{}l@{}}Semantic segmentation \cite{Wang2015}\\Depth estimation \cite{Laina2016}\end{tabular}                                        & \begin{tabular}[c]{@{}l@{}}Intel Xeon CPU, 2.4GHz\\Quadro K5200 GPU for CNN networks\end{tabular}                                                         \\ \hline         
    FLaME \cite{Greene2017}                                                   & 2017                           & \cmark                    & -                      & -                             & -                                & -                                & -                                 & -                             & \cmark               & -                       & -                                        & Intel Skull Canyon NUC (90 FPS) with \cite{Steiner2017}                                                         \\ \hline    
    DeepVO \cite{Wang2017}      & 2017      & \cmark        & -     & -     & -     & \multicolumn{2}{c}{End-to-end DL model}        & -     & -     & -     & \begin{tabular}[c]{@{}l@{}}Feature extraction\\LSTM pose estimation\end{tabular}       & Training on Tesla K40 GPU     \\ \hline    
    VINet \cite{Clark2017VINetVO}       & 2017      & \cmark        & -         & \cmark        & -     & \multicolumn{2}{c}{End-to-end DL model}        & -     & -     & -     & \begin{tabular}[c]{@{}l@{}}Features extraction (160ms)\\IMU LSTM (5ms) and Core LSTM\end{tabular}        & Training on Tesla K80 GPU     \\ \hline    
    Voxblox \cite{Oleynikova2016}                                                 & 2017                           & -                      & -                        & -                               & RGB-D                            & -                                & -                                 & -                             & -                 & \cmark                     & Semantic segmentation \cite{Grinvald2019}                      & i7 2.1 GHz CPU \cite{Oleynikova2020}                                                         \\ \hline        
    SVO \cite{forster2016svo}                                                     & 2016                           & \cmark                    & \cmark                      & \cmark                             & -                                  & Semi-direct                      & Graph                             & Sparse                             & -                 & -                       & -                                        & UP Board (40 FPS), ODROID XU4 (50 FPS) \cite{Delmerico2018}                                                         \\ \hline
    DSO \cite{Engel2016}                                                     & 2016                           & \cmark                    &                        & -                               & -                                  & Direct                           & Graph                             & Sparse                             & -                 & -                       & -                                        & i7-4910MQ CPU (7 FPS)                                                         \\ \hline
    \cite{Teixeira2016}                                       & 2016                           & \cmark                    & -                      & -                             & -                                & -                                & -                                 & -                             & \cmark               & -                       & -                                        & i7 4700MQ (around 143 FPS per KF with \cite{Leutenegger2013})                                                         \\ \hline    
    BundleFusion \cite{Dai2016}                                            & 2016                           & -                    & -                        & -                               & RGB-D                            & Det.+Des.                      & GPU-solver                        & -                             & -                 & \cmark                     & -                                        & Titan X GPU (around 36 FPS)                                                       \\ \hline    
    Chisel \cite{Klingensmith2015} & 2015 & \cmark & - & \cmark & RGB-D & Tracking & Filtering & - & - & \cmark & - & \begin{tabular}[c]{@{}l@{}}Tango tablet 4GB RAM, quadcore CPU, Tegra K1 GPU\\Tango mobile phone, 2GB RAM, quadcore CPU\end{tabular} \\ \hline
    ElasticFusion \cite{Whelan2015}                                           & 2015                           & -                    & -                        & -                               & RGB-D                            & Direct                           & Graph \cite{Sumner2007}                & Surfels    & -                 & -                       & Semantic segmentation \cite{McCormac2017}                      & Core i7-4930K,  GeForce GTX 780 Ti (32 FPS)                                                         \\ \hline     
    InfiniTAM \cite{Kahler2015}                                               & 2015                           & -                    & -                        & \cmark                             & RGB-D                            & Direct                           & ICP / \cite{Madgwick2011}              & -                             & -                 & \cmark                     & -                                        & \begin{tabular}[c]{@{}l@{}}Tegra K1 (47 FPS), iPad Air 2 (21 FPS)\\DE5 PCIe board (44 FPS), DE1 FPGA SoC (2 FPS) \cite{Gautier2019}\end{tabular}                                                         \\ \hline    
    ROVIO \cite{Bloesch2015}                                          & 2015                           & \cmark                    & \cmark                      & \cmark                             & -                                  & Direct                           & Filtering                         & Sparse                             & -                 & -                       & -                                        & UP Board, ODROID XU4 (22 FPS) \cite{Delmerico2018}                                                         \\ \hline        
    OKVIS \cite{Leutenegger2013}                                          & 2015                           & \cmark                    & \cmark                      & \cmark                             & -                                  & Det.+Des.                      & Graph                             & Sparse                             & -                 & -                       & -                                        & UP Board (11 FPS), ODROID XU4 (3 FPS) \cite{Delmerico2018}                                                         \\ \hline
    LSD-SLAM \cite{engel14eccv}                                                & 2014                           & \cmark                    & -                        & -                               & -                                  & Direct                           & Graph                             & Dense                             & -                 & -                       & -                                        & FPGA Zynq-7020 SoC (4.55 FPS) \cite{Boikos2016}                                                         \\ \hline    
    SLAM++ \cite{Salas2013}                                      & 2013                           & -                    & -                        & -                               & RGB-D                            & Direct                           & Graph                             & -                             & -                 & \cmark                     & -                                        & GPGPU implementation                                                        \\ \hline    
    KinectFusion \cite{Newcombe2011}                                            & 2011                           & -                    & -                        & -                               & RGB-D                            & Direct                           & ICP                               & -                             & -                 & \cmark                     & -                                        & Zynq UltraScale+ MPSoC ZCU102 (27.5 FPS) \cite{Gkeka2021}                                                         \\ \hline    
    PTAM \cite{klein07parallel}                                                    & 2007                           & \cmark                    & -                        & -                               & -                                  & Detection                        & LBA / GBA                         & Sparse                             & -                 & -                       & -                                        & ODROID XU4 (up to 12 FPS) \cite{PIRE201727}                                                         \\ \hline
    MonoSLAM \cite{Davison2007}                                                & 2007                           & \cmark                    & -                        & -                               & -                                  & Detection                        & Filtering                         & Sparse                             & -                 & -                       & -                                        & Intel Pentium M 1.60 GHz (53 FPS)                                                         \\ \hline
    \end{tabular}
    }
\end{sidewaystable*}
%\end{table*}

%% file: table_section3.tex
% \usepackage{multirow}
\begin{table*}[!t]
\renewcommand{\arraystretch}{1.3}
\caption{Performances of hardware implementation on embedded platforms. $(\ast)$ represents the back-end rate to update the states and the sparse 3D map \cite{Suleiman2019}, $(\ast\ast)$ corresponds to the measured time between the input image to the updated state, $(\dagger)$ indicates that the method failed on one or more sequences, which have not been included in the average result \cite{Delmerico2018}.}
\label{tab:table_section4}
\centering
\resizebox{\textwidth}{!}{
\begin{tabular}{lcclclc}
\hline
\multicolumn{1}{c}{\textbf{Methods}} & \textbf{Year} & \textbf{HW implementation} & \multicolumn{1}{c}{\textbf{Rate}}                       & \textbf{Power}                                           & \multicolumn{1}{c}{\textbf{ATE RMSE (m)}}                                   & \textbf{Dataset}        \\ \hline
CNN-SLAM proc. \cite{Li2019}                         & 2019          & \multirow{2}{*}[-1.5ex]{ASIC}                      & 80 FPS                                                   & \begin{tabular}[c]{@{}c@{}}243.6mW\\ 61.8mW\end{tabular} & \begin{tabular}[c]{@{}l@{}}97.90\% in tr.;\\ 99.34\% in rot.\end{tabular} & KITTI                   \\
Navion \cite{Suleiman2019}                                 & 2018          &                        & \begin{tabular}[c]{@{}l@{}}71 FPS\\ 19 FPS$^{\ast}$\end{tabular} & 24mW                                                     & 0.23                                                                     & EuRoC                   \\ \hline
VINS-Mono \cite{Qin2018}                              & 2018          & \multirow{6}{*}{ODROID}                           & 7 FPS$^{\ast\ast}$                                                   & \multirow{6}{*}{10W}                                                          & 0.16                                                                     & \multirow{6}{*}{EuRoC}                        \\
MSCKF-based \cite{Zhu2017}                                  & 2017          &                            & 20 FPS$^{\ast\ast}$                                              &                                                          & 0.56                                                                     &                         \\    
SVO+MSF \cite{forster2016svo}, \cite{Lynen2013}                                & 2016          &     & 50 FPS$^{\ast\ast}$                                                  &                                      & 0.69$^{\dagger}$                                                                     &  \\  
SVO+GTSAM \cite{forster2016svo}, \cite{Kaess2011}                              & 2016          &                            & 66 FPS$^{\ast\ast}$                                                  &                                                          & 0.11$^{\dagger}$                                                                     &                         \\  
ROVIO \cite{Bloesch2015}                                  & 2015          &                            & 22 FPS$^{\ast\ast}$                                                  &                                                          & 0.35                                                                     &                         \\  
OKVIS \cite{Leutenegger2013}                                  & 2013          &                            & 3 FPS$^{\ast\ast}$                                                   &                                                          & 0.26$^{\dagger}$                                                                     &                         \\ \hline
VINS-Mono \cite{Qin2018}                               & 2018          & \multirow{6}{*}{UP Board}                           & 7 FPS$^{\ast\ast}$                                                   & \multirow{6}{*}{15W}                                                         & 0.15                                                                     & \multirow{6}{*}{EuRoC}                        \\ 
MSCKF-based \cite{Zhu2017}                                  & 2017          &                            & 20 FPS$^{\ast\ast}$                                                  &                                                          & 0.53                                                                     &                         \\   
SVO+MSF \cite{forster2016svo}, \cite{Lynen2013}                                & 2016          &   & 40 FPS$^{\ast\ast}$                                                  &                                      & 0.69$^{\dagger}$                                                                     &                          \\  
SVO+GTSAM \cite{forster2016svo}, \cite{Kaess2011}                              & 2016          &                            & 50 FPS$^{\ast\ast}$                                                  &                                                          & 0.12$^{\dagger}$                                                                     &                         \\  
ROVIO \cite{Bloesch2015}                                  & 2015          &                            & -                                                        &                                                          & -                                                                        &                         \\  
OKVIS \cite{Leutenegger2013}                                  & 2013          &                            & 11 FPS$^{\ast\ast}$                                                  &                                                          & 0.27$^{\dagger}$                                                                     &                         \\ \hline
\end{tabular}
}
\end{table*}

%% file: overview_methods.tex
Table \ref{tab:table_slam} provides an overview of existing visual(-inertial) odometry, SLAM and 3D reconstruction methods. It highlights the strategies employed for the localization front-end (FE) and back-end (BE), the type of generated representation from cartography (point clouds and surfels) to volumetric reconstructions, the potential use of DNNs and the hardware (HW) implementation. The FPS measurements correspond to the performance of the entire pipeline.\par

The localization strategy shows that most methods employ a tracking front-end that refers to the KLT algorithm \cite{Bouguet1999PyramidalIO}. It allows a faster computation for local optimization compared to feature detection and description (Det.+Des). In \cite{Rosinol20icra-Kimera}, tracking takes an average of 4.5ms with 300 features per frame while Det.+Des takes around 15ms to extract 1200 ORB features per frame \cite{Campos2020}. However, the latter allows loop closures by the description of features. \par

Most of the methods are based on the graph back-end to exploit the sparsity of SLAM and to provide more accurate estimated trajectories \cite{Delmerico2018} compared to filtering techniques. The hardware implementations highlight the complexity of each approach for embedded platforms. For instance, the MSCKF-based \cite{Zhu2017} provides a higher performance in FPS than VINS-Fusion \cite{Qin2019}, which is based on a graph BE on UP Board with 20 FPS and 7 FPS respectively. However, the SVO graph-based method \cite{forster2016svo} runs at 40 FPS on the same platform. The front-end parameters, multithreading strategies, the sparsity of the map reconstruction used are important factors to take into account for real-time performance. As it can be seen, several 3D reconstruction methods \cite{Greene2017, Schps2020SurfelMeshingOS, Teixeira2016, Piazza2018, Oleynikova2016, Narita2019} do not provide a localization strategy (referred as: -), which means that they take advantage of 3D poses generated by SLAM methods. \par

Some approaches are based on deep learning in an end-to-end or hybrid configuration. At the moment, end-to-end pipelines do not achieve the performances of conventional methods in terms of accuracy of the estimated trajectory. For instance, the conventional monocular configuration \cite{Geiger2011} is outperformed by the supervised method \cite{Wang2017} but not the stereo configuration, which provide a $t_{rel}$ of 17.48\%, 5.96\% and 1.89\%. In \cite{Li2021}, the end-to-end unsupervised approach gives a $t_{rel}$ of 5.58\% compared to 3.21\% and 1.89\% for conventional approaches \cite{Mur-Artal2015, Geiger2011} respectively on testing sequences of the KITTI dataset \cite{Geiger2012CVPR}. However, hybrid pipelines give suitable solutions to include specific DNN-based functions. In \cite{Merrill2021ICRA}, the depth estimation from DNN \cite{Wofk2019} provides a processing time of 17.07ms and 7.09ms on the GPUs of Jetson Nano and TX2 respectively and 66.41ms and 105.07ms on the CPUs respectively with the Apache TVM optimization. On the same platforms, the block matching (BM) of the OpenCV library performs at 19.24ms, 12.38ms on GPUs and 27.76ms, 19.49ms on CPUs respectively. It demonstrates that the network prediction is faster on GPU with the TVM optimization, but the BM provides real-time performances on both implementations with lower processing times on CPUs. The accuracy of the dense conventional depth algorithms has not been quantified. However, the robustness of the deep learning-based estimation is demonstrated by a cleaner depth map than the noisy results of conventional methods \cite{Martins2018}. \par

%% file: intro_s3.tex
Localization and 3D reconstruction functions require a lot of hardware resources. The main HW implementations highlighted in table \ref{tab:table_slam} include high power and flexible CPUs/GPUs (i7 desktop/GTX), embedded CPUs/GPUs (Pi 3B+, ODROID, UP Board/TX2, Nano), and specific HW/SW co-design (FPGAs SoC). Transport systems, including autonomous cars to restricted MAVs, miniaturized robots, AR/VR systems, have a limited form factor and power budget. Those constraints affect the choice of implementation to meet the required accuracy and real-time performances. For instance, the power budget is around 10W-300W for autonomous cars \cite{liu2020computing}, 10W-15W for MAVs \cite{Delmerico2018, Skydio} and 10mW-10W for miniaturized robots and AR/VR devices \cite{Chatzopoulos2017, Suleiman2019, Palossi2019, Terry2019}. \par

Solutions and trade-offs for computing localization and 3D reconstruction functions in resource-constrained systems motivates the purpose of this section.

%% file: intro_hw.tex
A heterogeneous system is composed of various calculation resources. Components off-the-shelf (COTS), energy-efficient hardware accelerators or compact systems with low power consumption like vision chips are part of the available embedded platforms allowing the partitioning of advanced functions from sensors to 3D reconstruction. \par

%% file: table_fe_hw.tex
\begin{table}[!t]
\renewcommand{\arraystretch}{1.3}
\caption{Number of pixels processed per second for feature extraction functions based on SW and HW implementations.}
\label{tab:table_feathw}
\centering
\resizebox{\linewidth}{!}{
\begin{tabular}{lcccr}
\hline
\multicolumn{1}{c}{\textbf{HW implementation}} & \textbf{Features} & \textbf{Res.} & \textbf{FPS} & \multicolumn{1}{c}{\textbf{MP/s}} \\ \hline
Intel Core2Duo \cite{Nikolic2014}                                & Harris         & WVGA        & 40           & 14.44                            \\
Intel i7-4790 \cite{Qin2018} & Shi-Tomasi & WVGA & 66 & 23.82 \\
ARM Cortex-A15 \cite{Suleiman2019} & KLT Shi-Tomasi & WVGA & 19 & 6.86 \\
Jetson TX2 \cite{Rosinol21arxiv-Kimera} & KLT Shi-Tomasi & WVGA & 100 & 36.10 \\
FPGA Xilinx Zynq \cite{Nikolic2014}                          & Harris         &    WVGA     & 333          & 120.20                           \\
FPGA Xilinx Zynq \cite{Lepecq2020}                       & Harris+SURF    & VGA        & 320          & 98.30                           \\
ASIC VIO \cite{Suleiman2019} & KLT Shi-Tomasi & WVGA & 71 & 25.63 \\ 
ASIC CNN-VO \cite{Li2019} & CNN features & VGA & 80 & 24.58 \\ \hline
\end{tabular}
}
\end{table}

%% file: sensor_hw_cots.tex
In order to achieve real-time processing with limited resources, a trade-off between accuracy, robustness, execution time, memory management and power consumption must be reached \cite{Delmerico2018}. Several VIO methods \cite{forster2016svo, Bloesch2015, Qin2018, Leutenegger2013, Zhu2017} have been implemented on two hardware platforms for MAVs. It includes the UP Board with a quad-core Intel Atom x5-Z8350 1.44GHz CPU, 4Go RAM, a power consumption around 12W, and the ODROID XU4 with a hybrid ARM, a quad-core ARM A7 1.5GHz and an ARM big.LITTLE configuration quad-core A15 at 2.0GHz. ODROID has 2 GB RAM and a power consumption of 10W. \par

Table \ref{tab:table_section4} shows the real-time performances (in FPS) for each VIO method on COTS platforms. For instance, the graph-based VINS-Mono \cite{Qin2018} provides 7 FPS while the MSCKF-based \cite{Zhu2017} is at 20 FPS on ODROID compared to 20 FPS and 40 FPS respectively on the reference laptop. Real-time performance is achieved by reducing the number of features per frame, the size of the optimization sliding window and by including advanced single instruction multiple data (SIMD) instructions, like Intel SSE and ARM NEON optimizations for UP Board and ODROID respectively. The accuracy is not impacted by the optimizations performed for VINS-Mono with an average of the absolute translation error (ATE RMSE) of 0.16m, 0.16m, 0.15m for the laptop, UP Board and ODROID respectively. Although, the MSCKF-based is affected with an average ATE RMSE of 0.41m, 0.53m, 0.56m respectively. ROVIO \cite{Bloesch2015} is the only one that does not run on UP Board due to the CPU clock speed, which highlights the complexity of implementing VIO methods on different embedded platforms. \par

Partitioning a complete localization method with loop closures requires high computing resources. In \cite{Silveira2020}, ORB-SLAM2 \cite{Mur-Artal2017}, which integrates a loop closure module has been optimized using NEON instructions to take advantage of the advanced SIMD used in the ARM processors of the Raspberry Pi 3B+ and Jetson Nano. It achieves an average tracking time of 6.11 FPS on Raspberry Pi 3B+ and 9.64 FPS on Jetson Nano with input images at a 752$\times$480 resolution. This work focused on processing time and not on accuracy for the embedded implementations. The fact that the literature provides few information on the optimizations to be performed when using embedded platforms points to a line of research towards embedded SLAM, which incorporates loop closures. \par

A dense 3D representation is a challenge task to implement on embedded platforms due to the computation requirements. While the SVO method \cite{forster2016svo} implemented on ODROID U3 provides 3D poses, a W530 Lenovo laptop is used to compute a dense point cloud representation \cite{Pizzoli2014} on an NVIDIA Quadro K2000M GPU \cite{Faessler2015}. The 3D poses and input images are broadcasted at a frequency of 5Hz on a WiFi communication between the embedded platform and the laptop. Memory and speed efficient data structures tackle the limitations of a fixed-size volume and the large amount of memory required for volumetric reconstructions \cite{Newcombe2011}. A moving TSDF volume \cite{Whelan2012}, an octree-based approach \cite{hornung13auro, Steinbrucker2014} and a hashing scheme \cite{Niesner2013} enable the reconstruction of large scale environments with compact data structures. The hashing scheme has been used in several researches to save memory and CPU budget \cite{Oleynikova2016, Klingensmith2015, Muglikar2020} as it allows a complexity of $\mathcal{O}(1)$ compared to $\mathcal{O}(\log n)$ for octree structures \cite{hornung13auro}. \par

In order to provide more available resource in the main embedded system, partitioning the localization pipeline with custom hardware implementations is useful to cope with the high computational complexity of advanced algorithms, like feature extraction. \par

In \cite{Boikos2016}, specific units of the localization part of a semi-dense SLAM \cite{engel14eccv} has been accelerated on FPGA with a high-level synthesis (HLS) compiler. HLS compiler is used for performing low-level design optimizations based on conventional algorithms to increase the overall performance of the system. It achieves a framerate of 4.55 FPS compared to 2.27 FPS with a software-only implementation and a total power consumption of about 2.5W. In \cite{Nikolic2014}, the visual-inertial (VI) system consists of two imaging sensors with a resolution of 752$\times$480 (Aptina MT9V034) synchronized with an IMU (ADIS16488) through an ARM-FPGA Xilinx Zynq 7020 processing. The latter has been used to speed up the detection of Harris \cite{harris1988combined} and FAST \cite{Rosten2006} features in order to allocate more CPU resources for other tasks \cite{Nikolic2014}. Table \ref{tab:table_feathw} shows the performances of feature extraction methods on SW laptops CPU \cite{Qin2018, Nikolic2014}, Jetson TX2 \cite{Rosinol21arxiv-Kimera} and HW FPGAs \cite{Nikolic2014, Lepecq2020}, ASICs \cite{Suleiman2019, Li2019}. Software solutions are easier to program with more flexibility, but less efficient than dedicated HW FPGAs and ASICs. The number of megapixels processed per second (MP/s), which is related to the image resolution (Res.) of 752$\times$480 (WVGA) and 640$\times$480 (VGA) and the processing time (FPS), increases significantly with the implementation on FPGAs as the HW design is entirely dedicated to the feature extraction function. The application-specific integrated circuit (ASIC) VIO and ASIC CNN-VO are not only focused on this function. They are designed to integrate a full VO/VIO pipeline, which explains the difference in performance compared to FPGAs.

%% file: figure_compare_section.tex
\begin{figure*}[!t]
\centering
\includegraphics[width=\textwidth]{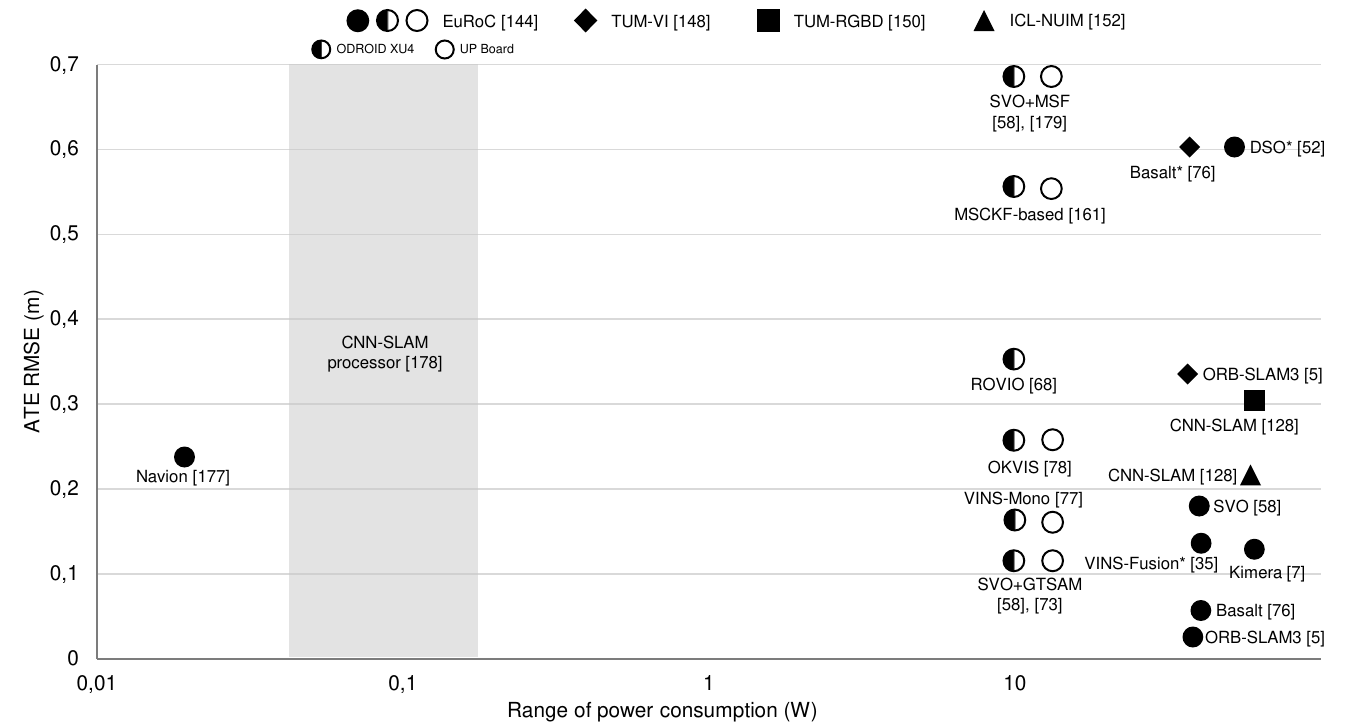}
\caption{Comparison of state-of-the-art methods in terms of accuracy and range of power consumption (W) between several hardware platforms, from ASIC to CPU-based. The vertical axis corresponds to the average RMSE in meters in all successful dataset sequences provided by the authors. Methods denoted by (*) means that the error has been obtained from \cite{Campos2020}.}
\label{fig:figure_compare_section}
\end{figure*}

%% file: sensor_hw_asi.tex
Specialized hardware, such as ASICs, give more freedom to design specific localization and 3D reconstruction functions. They offer high real-time performance and low power consumption. They are also more expensive in terms of development and fabrication \cite{Zhang2017}. \par

HoloLens2 \cite{Hololens} is one of the most advanced perception system based on the development of advanced functions to perceive an accurate topology of the environment with the generation of a mesh by leveraging multiple sensors with restrictive resources. It integrates a custom ASIC to perform the 6DoF localization and 3D mesh reconstruction functions, called the Holographic Processor Unit (HPU). The HPU processes input sensors data, including an IMU, a time-of-flight (ToF) depth sensor and four grayscale cameras at 30 FPS. Over all 160$\times$480 four channel resolutions, an image of 640$\times$480 is represented. The ASIC consumes less than 10W, can process more than 1 TOPS (Tera Operations Per Second) and contains 125 MB SRAM. It consists of 2 billion transistors in a 79mm$^2$ die size, 7 SIMD Fixed Point (SFP) for 2D processing, 6 Vector Floating Processor (VFP) for 3D processing and one dedicated core to DNNs processing programmable by Microsoft \cite{Terry2019}. \par

Navion is an energy-efficient VIO accelerator \cite{Suleiman2019}, which consists of three main parts: the Vision Front-End (VFE), the IMU Front-End (IFE) and the Back-End (BE) with local optimizations. Feature tracking (FT) is the only function in VFE to be performed per-frame. The remaining pipeline is based on the processing of keyframes. Based on input images at a 752$\times$480 resolution, the feature detection and tracking achieves an average of 71 FPS, which is comparable to other hardware platforms illustrated in table \ref{tab:table_feathw}. The performance is similar to software implementation on Intel i7. The ASIC provides efficient optimizations to reduce the memory usage. These include image compression in VFE, memory size reduction in BE and the way tracked features are stored in BE.
\iffalse
, such as programmable parameters to reduce power consumption according to the difficulty of the environment. For instance, small numbers are sufficient for sequences slow motions and bright scenes. These include the number of tracked features per frame and size of the sliding window in the BE. 
\fi 
The architecture optimizations reduce the initial memory size from 3.5MB to 854KB. To the best of our knowledge, the energy-efficient accelerator has not been publicly evaluated on a real MAV device, but on the Euroc dataset \cite{Burri2016}. With an average 71 FPS tracking process at framerate, the chip produces 19 FPS at keyframe rate and consumes around 24mW. In terms of energy-efficiency, Navion performs 0.43-2.5 TOPS/W. \par

Navion is entirely based on geometric methods. In \cite{Li2019}, the VO architecture is organized in three main parts. A CNN architecture to extract features from the input VGA image, a Perspective-n-Points (PnP) unit to compute the 2D-3D ($[R | t]$) 6DoF camera motion and a bundle adjustment (BA) unit to optimize on the last 20 keyframes. The CNN-based ASIC achieves an energy-efficiency of 3.6-5.34 TOPS/W, a latency of 12.5ms and consumes 243.6mW at 80 FPS VGA and reduces to 61.8 mW at 30 FPS VGA. \par

Partitioning the use loop closures for SLAM in a localization pipeline has been addressed by NeuroSLAM that integrates a SLAM architecture with a spiking neural network-based (SNN-based) \cite{Yoon2021}. While the VIO accelerators from the literature are designed with digital signals, NeuroSLAM additionally provides analogue signals to mimic SNNs. It achieves an energy-efficiency of 7.25-8.79 TOPS/W with a power consumption of 17.27-23.82mW respectively. Motivated by ultra-low power applications, the use of SNNs for SLAM opens new lines of research on this type of network. \par

Figure \ref{fig:figure_compare_section} illustrates ASIC implementations in the range of power consumption in mW. Several datasets are used to assess the accuracy of these implementations under real conditions. The accuracy of the method differs widely from one dataset to another. For instance, Navion \cite{Suleiman2019} has an average error of 0.23m in the Euroc sequences, while Basalt \cite{Usenko2020} provides an average error of 0.051m with the same sequences and 0.6m on handheld sequences with the TUM-VI dataset. The figure also exhibits the gap between the number of real-time methods developed with high resources, embedded COTS hardware and specific energy-efficient accelerators. \par

%% file: table_section3recons.tex
% \usepackage{multirow}
\begin{table*}[!t]
\renewcommand{\arraystretch}{1.3}
\caption{Implementation of 3D scene reconstruction methods on embedded platforms.}
\label{tab:table_section4recons}
\centering
\resizebox{\textwidth}{!}{
\begin{tabular}{lccrc}
\hline
\multicolumn{1}{c}{\textbf{Methods}} & \textbf{3D reconstruction}  & \textbf{HW implementation}               & \multicolumn{1}{c}{\textbf{Rate}} & \textbf{Use case}                 \\ \hline
FLaME \cite{Greene2017}                                  & Mesh                        & Intel Skull Canyon NUC i7 CPU flight computer   & 90 FPS                             & \multirow{2}{*}{MAV}              \\ 
Voxblox \cite{Oleynikova2016}                                & Volumetric & Asctec Firefly Intel i7 2.1 GHz CPU      & $>$4 FPS                 &                                   \\ \midrule[.1em] 
InfiniTAM \cite{Gautier2019}                              & \multirow{3}{*}{Volumetric}                             & Cyclone V Terasic DE1 FPGA SoC                     & 2 FPS                              & \multirow{3}{*}{Synthetic dataset \cite{handa:etal:ICRA2014}} \\ 
KinectFusion \cite{Gkeka2021}                           &                              & SoC FPGA Zynq UltraScale+ MPSoC ZCU102 & 27.5 FPS                           &                                   \\ 
InfiniTAM \cite{Gautier2019}                              &                              & Stratix V Terasic DE5 PCIe board                   & 44 FPS                             &                                   \\ \hline
\end{tabular}
}
\end{table*}

%% file: sensor_hw_vchip.tex
Vision chips offer a specific implementation for integrating complex algorithms for a wide range of applications requiring low latency image processing \cite{Millet2019, Dudek2005}. In addition to the imaging sensor, processing units are integrated for in-sensor computing, which increases the overall performance of the system. Vision chips have the capacity to compute feature extraction functions at very high speed in a compact and ultra low power system. In \cite{Murai2020}, the vision system is designed to extract FAST features \cite{Chen2017FeatureEU} and describe them with a 44-bit binary-edge descriptor. This system operates at 300 FPS. The remaining part of the VO pipeline runs on an Intel i7-6700HQ CPU with binary edges and corners images that are tolerant of motion blur compared to conventional visible cameras.

%% file: intro_algo_hw.tex
This subsection provides an overview of algorithmic approaches used for several embedded systems from COTS platforms to specific HW/SW co-design implementations. The localization part is dissociated from the 3D reconstruction to obtain a broader view of the current implementations.

%% file: embed_algo_slam.tex
HoloLens \cite{Hololens} enables spatial mapping with mesh generation, spatial processing for finding planes and spatial understanding with semantic labels. The HPU integrates a localization pipeline \cite{Ebstyne2016} to provide an accurate pose estimation. Based on the use of inertial measurements and imaging sensors, a block filter consisting of an EKF and a sensor fusion algorithm is developed. On a closed loop path of 287m, the HoloLens localization system drifts 2.39m from start to finish \cite{Huebner2020} \cite{Khoshelham2019}. The overall quality of the 3D model provided by the device has some holes in the mesh. Its accuracy has been measured from a ground truth point cloud. The reconstruction of several offices provided by the AR/VR device gives an average Euclidean distance of 0.023m with a fixed scale between the 3D reconstruction and the ground truth. \par

The localization strategy implemented on the HPU gives a large view of the type of algorithms to be implemented that provide real-time pose estimations. The autonomous quadrotor system \cite{Faessler2015} highlights the localization parameters to limit the resource usage. The SVO visual odometry method \cite{forster2016svo} implemented in ODROID U3 uses two threads to estimate the camera motion and to insert keyframes into the extended map. The \textit{fast} parameter of the approach has been used, which limits to 120 the number of detected features per frame and keeps in memory a maximum of 10 generated keyframes in the map. In order to obtain a robust system, the IMU data and the poses are merged via the MSF method \cite{Lynen2013}, which uses an EKF. The experiments were carried out on a 20m long, 1.7m high indoor path and on a 100m and 20m outdoor path respectively. The system comprising SVO+MSF achieves a maximum drift of 0.5\% of the travelled distance and an average trajectory error (ATE RMSE) of about 0.05m for a closed loop trajectory. \par

%% file: embed_algo_recons.tex
Table \ref{tab:table_section4recons} illustrates that only few methods from mesh to voxel-based have been used onboard MAVs or by taking advantage of HW/SW co-design on FPGA/SoC. \par

The FLaME mesh reconstruction method \cite{Greene2017} has been implemented on an Intel Skull Canyon NUC flight computer. The MAV was equipped with a Point Grey Flea 3 camera operating at 60 FPS with an image resolution of 320$\times$256 in addition to an IMU. The experiments were conducted in indoor and outdoor environments with a vehicle speed of 2.5$m/s$ and 3.5$m/s$ respectively. It reconstructed a very detailed mesh representation at a framerate over 90 FPS. The same use case has been used for the Voxblox volumetric reconstruction \cite{Oleynikova2016, Oleynikova2020}. The ROVIO odometry method \cite{Bloesch2015} provides pose estimations to the reconstruction approach. Voxblox has been evaluated on a MAV platform equipped with an Intel i7 2.1 GHz CPU and a stereo camera synchronized to an IMU. With a voxel size representation of 0.2m, the computation time of the complete system is less than 250ms. In order to obtain an accurate scene representation with a lower granularity, the use of HW optimizations is particularly useful for maintaining real-time performance. \par

Table \ref{tab:table_section4recons} shows several voxel-based reconstructions performed on HW FPGA/SoC. In \cite{Gkeka2021}, KinectFusion \cite{Newcombe2011} has been optimally implemented on a SoC FPGA Xilinx UltraScale+ MPSoC ZCU102. Raycasting is the only part of the pipeline to be computed on the ARM CPU due to complex memory access. The reconstruction method achieves a performance of 27.5 FPS with 320$\times$240 input images from the ICL-NUIM dataset. The optimizations focused on parameters that improve the execution time, so that the accuracy error of the method increases from around 0.018m to 0.08m from one sequence to the next. In \cite{Gautier2019}, InfiniTAM \cite{Kahler2015} has been implemented on a low-cost Terasic DE1 FPGA SoC and on a high-cost Terasic DE5 PCIe board. The real-time performance widely differs depending on the available resources, with a performance of 2 FPS and 44 FPS respectively with input depth images at a 320$\times$240 resolution. \par

Although FPGA SoC allows the acceleration of advanced functions on the hardware with the use of a CPU for other computations. The volumetric method \cite{Oleynikova2020} offers a CPU-based implementation, which provides research directions for implementing complex tasks for more accurate scene perception in a heterogeneous system.

%% file: conclusion.tex
In this paper, we have reviewed visual(-inertial) SLAM methods, from real-time localization with scene cartography to volumetric reconstruction in the context of resource-constrained embedded platforms. It highlights the different strategies employed for localization and reconstruction functions, including the potential use of deep neural networks. The latter is particularly useful in a hybrid configuration to combine the strengths of deep learning and the maturity of model-based methods for specific functions, including feature detection, description, matching and intermediate representations. This study also provides an overview of the hardware implementation of localization and reconstruction functions from COTS systems to specific ASIC/SoC integration. It shows the gap between algorithmic methods developed with the high resources available in conventional laptops and those developed for transport systems with limited resources, including MAVs, miniaturized robots and mobile AR/VR devices. \par

Several odometry methods are developed with limited algorithmic complexity and provide parameters to be configured for real-time processing on restrictive platforms. This survey shows that few SLAM and volumetric methods are developed in this specific context. The implementation of loop closures capability for SLAM remains a challenge to integrate due to the required computational resources. As real-time processing, memory management and low power consumption are essential, several questions remain open to find the best trade-off between accuracy, robustness, scalability and resource consumption. The required precision of intermediate representations, including depth estimation and semantic segmentation for an accurate 3D model, is one of the research areas to be explored, as well as the computational cost and use of hardware resources, especially with deep learning methods. The granularity of the 3D reconstruction also raises several questions. For instance, what is the required granularity or space limitation in the reconstruction? What is the right trade-off for accurate real-time localization and reconstruction with limited available resources?